\pdfoutput=1
\documentclass[10pt,twocolumn,letterpaper]{article}

\usepackage[pagenumbers]{cvpr}

\usepackage{multirow}
\usepackage{needspace}
\usepackage{pifont}

\newcommand{\ours}{LayoutRecall\xspace}
\newcommand{\data}{LayoutLoop\xspace}
\newcommand{\bsc}{BLC-Fourier\xspace}
\newcommand{\bslm}{BSLM\xspace}
\newcommand{\gamc}{GAMC\xspace}
\newcommand{\caloss}{CAL\xspace}

\newcommand{\figslide}[3]{\includegraphics[page=#1,trim=#2,clip,width=#3]{figures/figures_paper.pdf}}

\definecolor{cvprblue}{rgb}{0.21,0.49,0.74}
\usepackage[pagebackref,breaklinks,colorlinks,allcolors=cvprblue]{hyperref}

\title{\ours: Preserving Appearance in Multi-Round 3D Layout-Controlled Image Editing}

\author{Yankuan Chi$^{1}$ \quad
Xin Lin$^{1\dagger}$ \quad
Baihan Yang$^{1}$ \quad
Yuheng Liu$^{2}$ \quad
Jason Li$^{3}$ \quad
Truong Nguyen$^{1\ddagger}$\\[0.3em]
$^{1}$University of California San Diego \quad
$^{2}$University of California Irvine \\
$^{3}$Nanyang Technological University\\
}

\begin{document}

\twocolumn[{%
\renewcommand\twocolumn[1][]{#1}%
\maketitle
\begin{center}
  \centering
  \captionsetup{type=figure}
  \includegraphics[width=\linewidth,trim={5bp 288bp 60.5bp 5bp},clip]{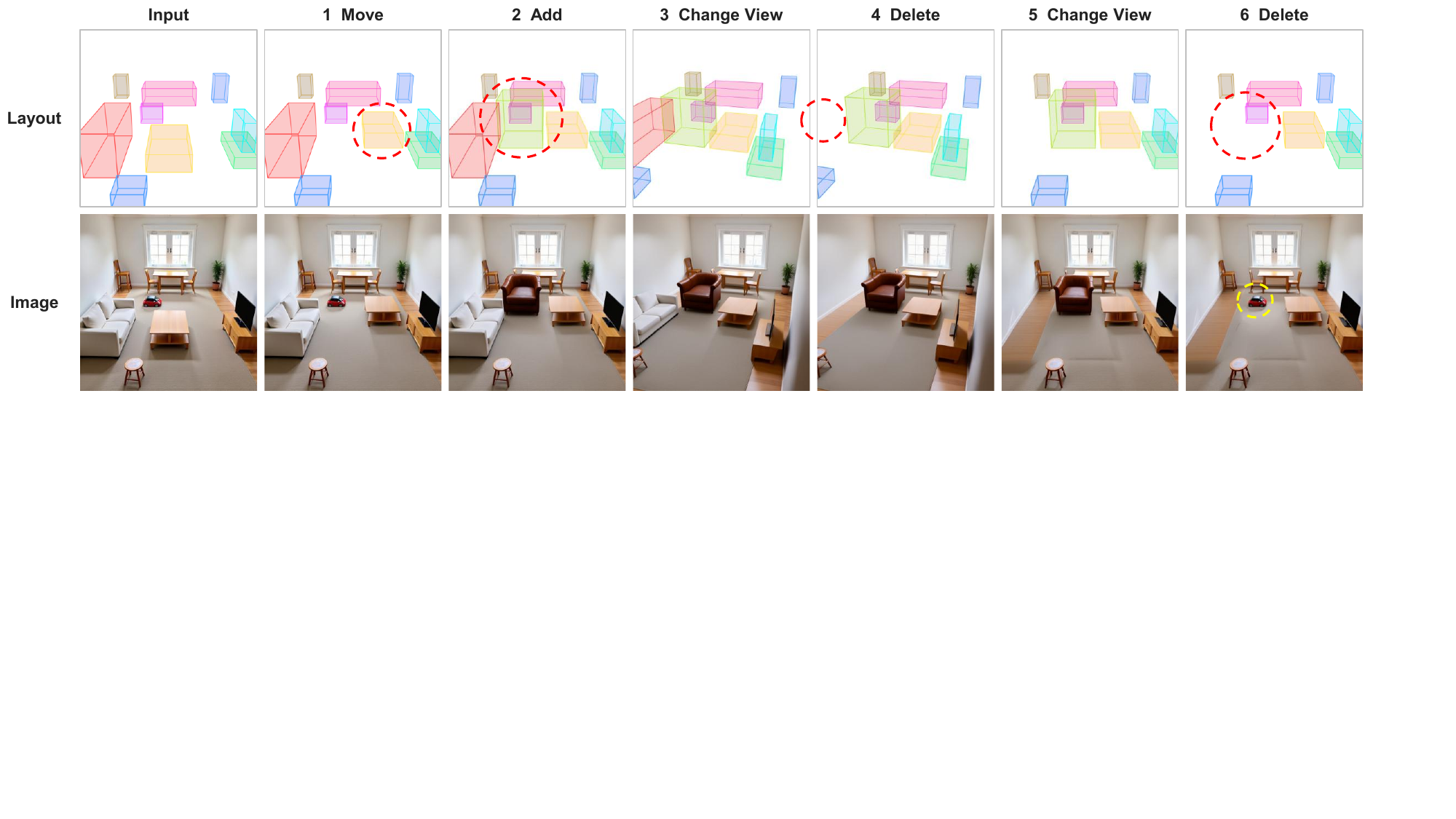}
  \captionof{figure}{\textcolor{black}{\textbf{Multi-round 3D layout-controlled image editing with \ours.}
  The user specifies the layout of the objects with 3D boxes (top) and edits the image (bottom) by adding, moving, or deleting an object box or changing the camera view in each round. \ours preserves appearance when an object reappears in a later round (toy car, yellow circle).}}
  \label{fig:teaser}
\end{center}%
\vspace{1em}
}]

\begingroup
\renewcommand{\thefootnote}{$\dagger$}%
\footnotetext{Project lead.}%
\renewcommand{\thefootnote}{$\ddagger$}%
\footnotetext{Corresponding author.}%
\endgroup

\begin{abstract}
\begingroup\color{black}
3D layout-controlled image editing gives users explicit control over object placement and camera viewpoint. In a sequence of edits, however, the latest image can lose the appearance of an object when it is occluded or leaves view. The target layout specifies where the object should reappear, while its appearance must be recovered from earlier observations. We introduce \ours, a multi-round editor that stores and reuses appearance in the local coordinate systems of 3D boxes. Box-Surface Latent Memory retains image latents on box faces and retrieves them along rays in the target view. Geometry-Aligned Memory Conditioning supplies the retrieved tokens to a FLUX.2 editor using target-view positions, shared box-local embeddings, and object-wise attention. A correspondence loss encourages the editor to attend to matching observations during denoising. We also introduce \data, a multi-round 3D layout-controlled image editing benchmark with calibrated layouts and designed revisits. We conduct extensive experiments on \data and real-scene data to verify the ability of our method. The project page is at \url{https://ykchizs.github.io/layoutrecall.github.io/}.
\par\endgroup
\end{abstract}

\Needspace{4\baselineskip}
\section{Introduction}
\label{sec:intro}
\begingroup\color{black}

3D layout-controlled generation and editing allow users to compose a scene by specifying object positions, orientations, and sizes with 3D boxes and adjusting the camera viewpoint~\cite{buildascene,laconic,seethrough3d,scenedesigner,boxctrl,tib}. This interface supports object addition, movement, and deletion with coarse geometry, making it useful for product visualization, scene design, and data generation for perception~\cite{mobi}. In practice, users refine scenes through successive edits, requiring precise spatial control and stable appearance.

Existing layout-conditioned editors control individual transformations from a source image~\cite{boxctrl,tib}. Reusing only the latest output loses evidence about objects that become occluded or leave view. A subsequent layout may bring an object back at a different position, but the box describes its geometry and leaves its original appearance underdetermined. The editor can therefore satisfy the requested placement while generating a different instance. This failure differs from the gradual image degradation observed in recursive editing~\cite{mtopsd,freqedit,anchoredit}. It also affects background regions exposed after an occluder moves away. Recovery requires retaining earlier observations and aligning them with the target after objects and cameras move.

We build \ours on the observation that a 3D box provides both a spatial control signal and a coordinate system that moves with the object. Box-Surface Latent Memory (\bslm) uses this coordinate system to retain observed appearance on the box surface and retrieve it through rays from the target view. This operation transfers a stored observation to the corresponding location in the requested layout. Geometry-Aligned Memory Conditioning (\gamc) then places the retrieved latents in the target image, embeds their box-local surface coordinates, and restricts attention to the appropriate object region. A Correspondence Attention Loss (\caloss) trains the FLUX.2 editor to use matching observations during denoising. Repeated reading and writing carries appearance forward through successive changes in layout and visibility (\cref{fig:teaser}).

We introduce \data to evaluate this capability on calibrated multiview sequences with object revisits and camera returns. A ten-round rollout measures image quality, editing accuracy, preservation, and consistency across editing rounds. We conduct extensive experiments on \data and real-scene data to evaluate these aspects.

Our contributions are as follows.
\begin{itemize}
  \item We develop \ours for multi-round 3D layout-controlled image editing, covering object addition, movement, deletion, and camera view change.
  \item We introduce \bslm and \gamc to store and retrieve appearance in object-local coordinates and guide denoising with spatial and semantic correspondences. \caloss supervises attention to matching observations.
  \item We construct \data, a calibrated multiview dataset and benchmark that uses revisits to test the overall image quality, editing accuracy, appearance preservation, and camera-return consistency.
\end{itemize}
\par\endgroup

\section{Related Work}
\label{sec:related}
\begingroup\color{black}

\noindent\textbf{3D layout control.}
Spatial conditions provide a means to control scene geometry in image synthesis and editing. Depth maps constrain dense scene structure~\cite{loosecontrol,buildascene}, while orientation and occlusion cues target specific relations~\cite{origen,compasscontrol,viewpointtokens,Cobl,DAEdit,Larender,Vodiff}. Another line of work utilizes explicit 3D representations as conditions, but at the cost of high computational demand and imperfect reconstruction~\cite{3dFixup,imageSculpting,objectMorpher,diffusionHandles,diffusionModelsareGeometryCritics,phyedit}. Three-dimensional boxes offer explicit object-level control. Neural Assets~\cite{neuralassets} and LACONIC~\cite{laconic} encode box parameters as tokens, and other approaches use boxes to guide object inpainting~\cite{mobi,omran2025placement}. SceneCraft~\cite{scenecraft}, SceneDesigner~\cite{scenedesigner}, SeeThrough3D~\cite{seethrough3d}, and Mise-en-Video~\cite{miseenvideo} condition generation on projected layouts. BoxCtrl~\cite{boxctrl} and Thinking-in-Boxes~\cite{tib} adapt related spatial interfaces to single-round image editing. Our work uses projected layouts to specify each edit and the underlying boxes to associate appearance across rounds.

\vspace{1mm}

\noindent\textbf{Multi-turn editing.}
Repeated edits accumulate errors as each generated output becomes the next input. Training-based methods adapt editors to imperfect inputs~\cite{acedit,editr2,emuedit,mtforcing,mteditflow,mtopsd,vincie}. Inference-time methods change frequency content, sampling, or model components to improve stability~\cite{freqedit,mcie,reedvae,vaelfa,emilie}. The above methods all focus on holistic image quality instead of each entity's appearance. Agents and history-based approaches can also condition on earlier turns~\cite{imagagent,anchoredit,mtedit,ExCave}. We study a complementary recovery problem in which a spatial edit removes visual evidence from the current frame and a later edit makes that content visible again. The target geometry provides a way to locate the relevant observations in the new view.

\vspace{1mm}

\noindent\textbf{Appearance memory for generation.}
Long video generators retain earlier observations through frame context, compressed history, or spatial retrieval~\cite{contextasmemory,framepack,mamba_vod,rmem,spatialmemory}. Entity-centric memories organize past appearances into slots or object groups~\cite{videomemory,slotmem,slotmemory,iamflow,talkingmemory,ocd,recapforcing,memorysurvey}. For explicit layout edits, an object's identity determines which history to retrieve, and its new box pose determines where that history should appear. \bslm addresses observations by both object identity and box-surface position. \gamc uses those addresses to place observations in the target view and condition the appropriate object regions.
\par\endgroup

\section{Method}
\label{sec:method}
\begingroup\color{black}

\subsection{Framework Overview}
\label{sec:overview}

\begin{figure*}[t]
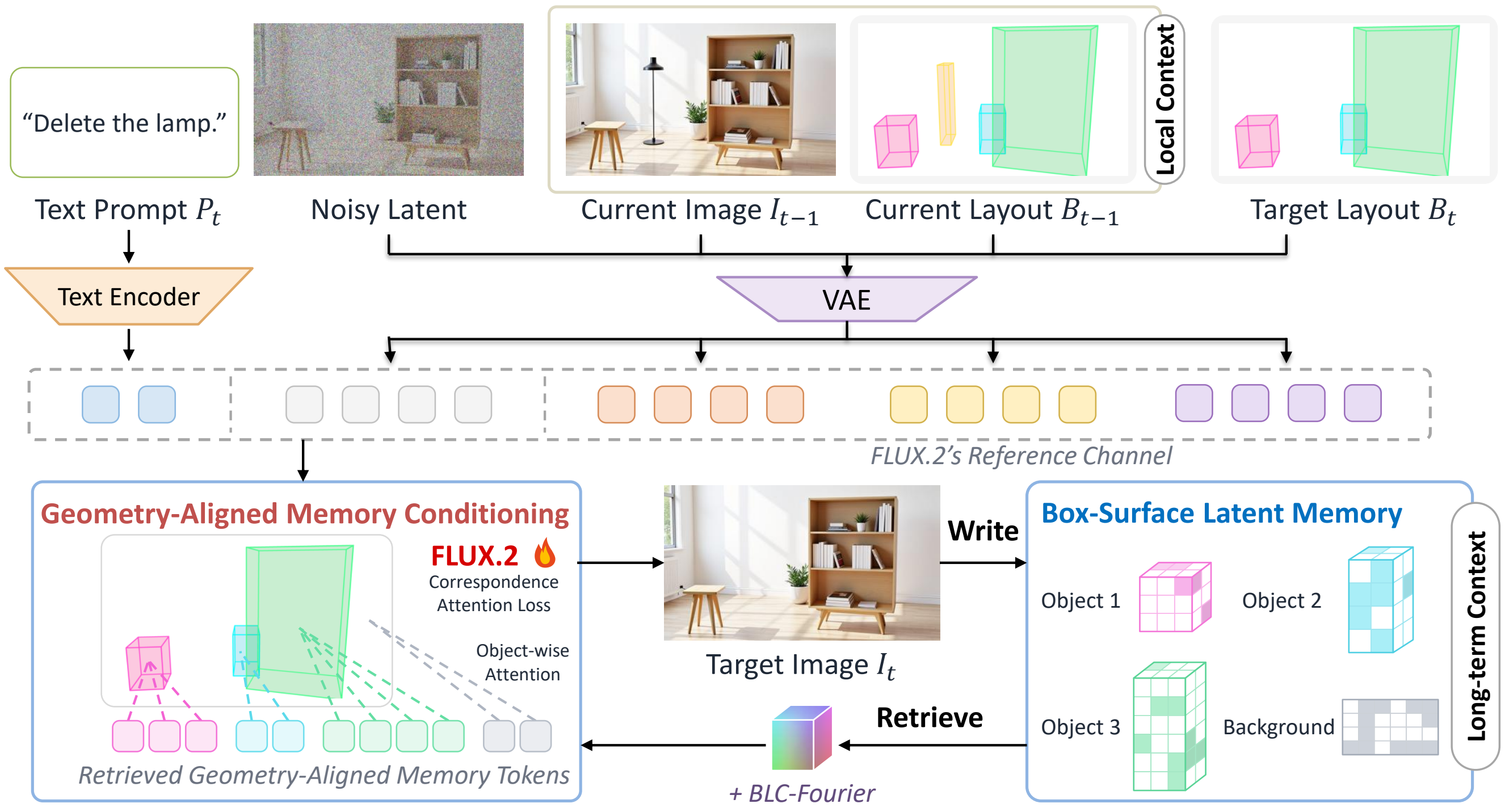

  \centering
  \figslide{1}{3 128.5 189.5 2}{\linewidth}
  \caption{\textbf{Overview of \ours.}
  In each editing round $t$, our model generates the target image $I_t$ that follows the target layout $B_t$. The current image $I_{t-1}$ and its layout $B_{t-1}$ serve as local context, while the Box-Surface Latent Memory (BSLM) provides long-term context. Relevant memory is retrieved from BSLM and injected into the FLUX.2 editor through Geometry-Aligned Memory Conditioning (GAMC). Once $I_t$ is generated, its observations are written back into BSLM.}
  \label{fig:overview}
\end{figure*}

Given an initial image $I_0$ and its 3D layout $B_0$, \ours generates an image after each requested edit (\cref{fig:overview}). In round $t$, the user adds, moves, or removes an object box, or changes the camera view. We denote the resulting boxes and camera calibration by $\Gamma_t$ and their rendered layout image by $B_t$. The edit prompt $P_t$, current image $I_{t-1}$, current layout $B_{t-1}$, and the target layout $B_t$ specify the current transformation. A persistent memory state $\mathcal{M}_{t-1}$ provides appearance evidence from earlier rounds. Starting with observations from $I_0$, we alternate retrieval from the memory, generation with the editor $G_\theta$, and memory update:
\begin{equation}
\begin{aligned}
  \mathcal{R}_t &= \operatorname{Retrieve}(\mathcal{M}_{t-1},\Gamma_t),\\
  I_t &= G_\theta(P_t,I_{t-1},B_{t-1},B_t,\mathcal{R}_t),\\
  \mathcal{M}_t &= \operatorname{Write}(\mathcal{M}_{t-1},I_t,\Gamma_t).
\end{aligned}
\label{eq:round}
\end{equation}
Box-Surface Latent Memory (\bslm, \cref{sec:memory}) defines the retrieval and writing operations. Geometry-Aligned Memory Conditioning (\gamc, \cref{sec:editing}) integrates the retrieved set $\mathcal{R}_t$ into the FLUX.2 editor $G_\theta$. A Correspondence Attention Loss (\caloss, \cref{sec:training}) trains the editor to attend to matching observations.

\subsection{Box-Surface Latent Memory}
\label{sec:memory}

\begin{figure}[t]
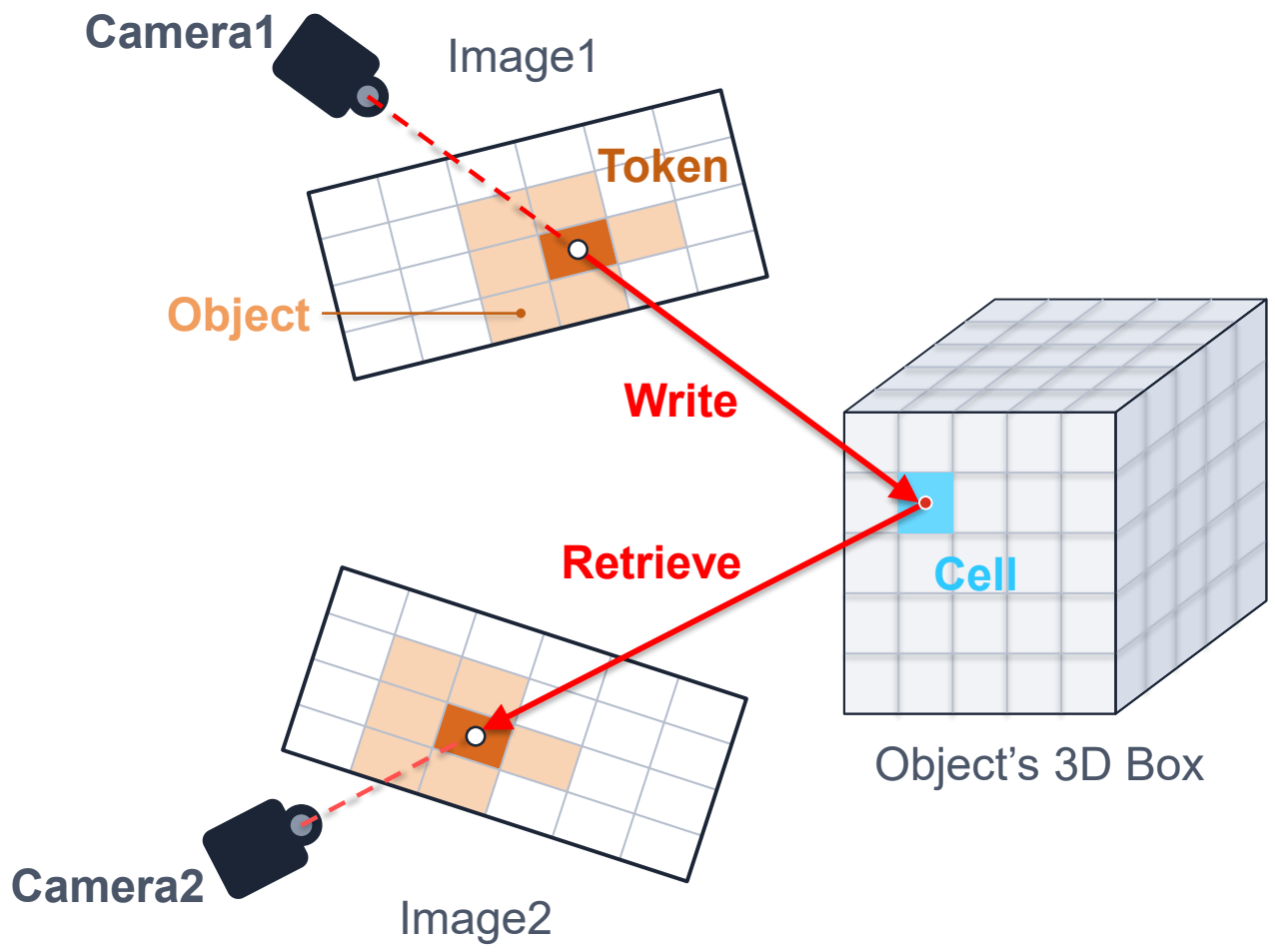

  \centering
  \figslide{2}{2.5 171.5 467 4.5}{\linewidth}
  \caption{\textcolor{black}{\textbf{Writing and retrieval in \bslm.} A ray from the camera through an image token's center maps it to a cell on its object's 3D box surface. Retrieval is performed in the same way with the target camera and box layout.}}
  \label{fig:memory}
\end{figure}

\Cref{fig:memory} illustrates how a camera ray maps an image token to a cell on the object's 3D box.

\vspace{1mm}

\noindent\textbf{Object-local addressing.}
A previously observed patch can move to a different image location after either the object or the camera moves. We therefore address appearance by object identity and location on the object's box to maintain both semantic and spatial correspondences. For a surface hit $\mathbf{x}$ on box $k$ with center $\mathbf{c}_k$, rotation $R_k$, and side lengths $\mathbf{s}_k$, the normalized local coordinate is
\begin{equation}
  \mathbf{u}_k(\mathbf{x})=\operatorname{diag}(2/\mathbf{s}_k)R_k^\top(\mathbf{x}-\mathbf{c}_k).
  \label{eq:boxlocal}
\end{equation}
Each face of the box is divided into cells of fixed physical size. A cell key consists of the object identity, face, and two grid indices. Each occupied cell stores a clean FLUX.2 VAE token and its box-local surface coordinate. The box provides a geometric proxy whose local coordinates preserve memory addresses across edits.

\begin{table*}[t]
  \centering
  \small
  \setlength{\tabcolsep}{4pt}
\caption{\textcolor{black}{\textbf{Single-round editing on \data.} Editing accuracy uses a separate metric for each operation: CLIP-T~\cite{clip} for addition, DINOv2~\cite{dinov2} similarity between source and target crops for movement, DINOv2 similarity between the output and the reference within the projected box of the removed object for deletion, and full-image DINOv2 similarity to the reference for camera changes. Best results are in \textbf{bold} and second best are \underline{underlined}.}}
  \label{tab:e1}
  \begin{tabular*}{\linewidth}{@{\extracolsep{\fill}}l c cccc ccc@{}}
    \toprule
    & \multicolumn{1}{c}{Quality} & \multicolumn{4}{c}{Editing Accuracy} & \multicolumn{3}{c}{Preservation} \\
    \cmidrule(lr){2-2}\cmidrule(lr){3-6}\cmidrule(lr){7-9}
    Method & $|\Delta$MUSIQ$|\downarrow$ & Add$\uparrow$ & Move$\uparrow$ & Delete$\uparrow$ & Camera$\uparrow$ & PSNR$\uparrow$ & SSIM$\uparrow$ & LPIPS$\downarrow$ \\
    \midrule
    BoxCtrl~\cite{boxctrl} & 10.93 & 0.203 & 0.500 & 0.576 & 0.643 & 22.37 & 0.801 & 0.301 \\
    Thinking in Boxes~\cite{tib} & \underline{3.27} & 0.215 & \underline{0.585} & \underline{0.691} & \underline{0.756} & \underline{26.03} & \underline{0.862} & \underline{0.202} \\
    SeeThrough3D~\cite{seethrough3d} & 7.57 & \textbf{0.234} & 0.308 & 0.261 & 0.270 & 9.32 & 0.418 & 0.803 \\
    \ours (ours) & \textbf{1.79} & \underline{0.218} & \textbf{0.602} & \textbf{0.774} & \textbf{0.792} & \textbf{32.66} & \textbf{0.911} & \textbf{0.132} \\
    \bottomrule
  \end{tabular*}
\end{table*}

\begin{table*}[t]
  \centering
  \small
  \setlength{\tabcolsep}{4pt}
  \caption{\textbf{Single-round editing on real scenes.} All metrics are reference-free. We measure the editing accuracy by the mIoU with the projected target boxes and the edit success rate detected by Grounding DINO~\cite{groundingdino}. Survive measures the fraction of unedited and visible objects that Grounding DINO still detects inside their boxes after the edit. Best results are in \textbf{bold} and second best are \underline{underlined}.}
  \label{tab:real_sr}
  \begin{tabular*}{\linewidth}{@{\extracolsep{\fill}}l c cc cccc@{}}
    \toprule
    & \multicolumn{1}{c}{Quality} & \multicolumn{2}{c}{Editing Accuracy} & \multicolumn{4}{c}{Preservation} \\
    \cmidrule(lr){2-2}\cmidrule(lr){3-4}\cmidrule(lr){5-8}
    Method & $|\Delta$MUSIQ$|\downarrow$ & mIoU$\uparrow$ & Succ.$\uparrow$ & Survive$\uparrow$ & PSNR$\uparrow$ & SSIM$\uparrow$ & LPIPS$\downarrow$ \\
    \midrule
    BoxCtrl~\cite{boxctrl} & 6.43 & \underline{0.712} & \underline{0.887} & 0.326 & 21.48 & 0.736 & 0.205 \\
    Thinking in Boxes~\cite{tib} & \underline{2.25} & 0.541 & 0.713 & \underline{0.742} & \underline{23.66} & \underline{0.764} & \underline{0.134} \\
    SeeThrough3D~\cite{seethrough3d} & 5.81 & 0.507 & 0.713 & 0.616 & 9.37 & 0.206 & 0.746 \\
    \ours (ours) & \textbf{1.57} & \textbf{0.728} & \textbf{0.900} & \textbf{0.916} & \textbf{29.33} & \textbf{0.900} & \textbf{0.047} \\
    \bottomrule
  \end{tabular*}
\end{table*}

\vspace{1mm}

\noindent\textbf{Writing visible appearance.}
We encode the current image using the FLUX.2 VAE and associate its tokens with visible objects. SAM~3~\cite{sam3} uses the object's description and its projected box to obtain a mask. A token is assigned to the object when mask coverage exceeds $\tau_{\mathrm{obj}}$. For each assigned token, we intersect the ray from the camera origin through the token's center with that object's box and write the token to the corresponding surface cell. A cell holds one token. We assess the token's image resolution and replace a stored token only when the new token improves the resolution by a certain margin. This rule limits repeated writes of coarser appearances as the camera and objects move. Background tokens are stored separately using the camera view and image location as the address.

\vspace{1mm}

\noindent\textbf{Retrieval in the target view.}
We cast rays through target tokens using the camera and boxes in $\Gamma_t$. Each object hit supplies a box surface address. If the intersected cell is empty, we look for the nearest occupied cell within a fixed radius on the same face. This neighborhood lookup accounts for sparse sampling of the box surface. Small changes in placement or viewpoint can shift target rays between previously observed cells. The retrieved tokens are placed at the projections of their stored surface locations under the target box and camera. Background retrieval uses the record for the target camera view when one is available.

\subsection{Geometry-Aligned Memory Conditioning}
\label{sec:editing}

The retrieval step identifies appearance evidence for the requested scene geometry. \gamc supplies that evidence to the editor with its target image position, object-local surface coordinate, and object identity. These three signals locate the observation in the output, indicate which surface it describes, and determine which object can access it.

\vspace{1mm}

\noindent\textbf{Target-view token placement.}
We fine-tune the FLUX.2~\cite{flux2} editor with operation-specific LoRA adapters~\cite{lora}. The VAE encodes $I_{t-1}$, $B_{t-1}$, and $B_t$ as clean reference tokens, which are appended to the noisy target tokens. Distinct time-axis offsets in RoPE~\cite{rope} identify the three reference streams (10, 20, and 30). We append retrieved memory tokens as a fourth stream at offset 40. Its spatial RoPE indices are the projected locations in the target image, so the positional encoding follows the requested placement and viewpoint.

\vspace{1mm}

\noindent\textbf{Box-local Fourier embedding.}
Target image coordinates place a memory token in the output image plane. To encode its box surface correspondence across views, we also provide the box-local surface coordinate $\mathbf{u}$ of the memory token, following the use of local box coordinates in SceneDesigner~\cite{scenedesigner}. We map $\mathbf{u}$ through Fourier features~\cite{fourierfeatures,nerf} and an MLP.
\begin{equation}
\begin{aligned}
  e(\mathbf{u}) &= \mathrm{MLP}\!\left([\mathbf{u},\phi(\mathbf{u})]\right),\\
  \phi(\mathbf{u}) &= \bigl[\sin(2^\ell\pi\mathbf{u}),
                       \cos(2^\ell\pi\mathbf{u})\bigr]_{\ell=0}^{5}.
\end{aligned}
\label{eq:bscfourier}
\end{equation}
We call this shared embedding \bsc. We add it to object tokens in the retrieved memory, previous image, and noisy target. The same box-surface point thus receives the same embedding across these streams. Background memory tokens instead receive a learned embedding $e_{\mathrm{bg}}$.

\vspace{1mm}

\noindent\textbf{Object-wise attention.}
Different objects can have identical local coordinates. We use the object identity attached to each retrieved memory token to mask access to the noisy target tokens to establish correct semantic correspondences. An object's memory tokens can only be attended to by the object's noisy target tokens. This mask pairs each region with its own appearance evidence, and \bsc provides the finer surface correspondence within the region.

\subsection{Correspondence-Supervised Training}
\label{sec:training}

\noindent\textbf{Correspondence Attention Loss.}
Our preliminary runs assigned little attention to matched target and memory tokens even when their semantic and geometric correspondences were available. Thus, we train the editor to use these matches with \caloss. For a target token $z$ of object $k$, let $S_z$ contain the object's tokens from $I_{t-1}$ and the retrieved memory that belong to $k$ and lie within $\rho$ token footprints of $z$'s surface point. Let $a_z$ be the attention mass assigned to $z$ over $S_z$, and let $\mathcal{Z}$ be the set of target object tokens with at least one match. We apply a hinge penalty below the threshold $\beta$.
\begin{equation}
  \mathcal{L}_{\mathrm{corr}} = \frac{1}{|\mathcal{Z}|}
    \sum_{z\in\mathcal{Z}}\max\!\left(0,\log\beta-\log a_z\right).
\label{eq:corr}
\end{equation}

\noindent\textbf{Training on short revisits.}
We train on two-edit sequences $(I_0,I_1,I_2)$ in which the final round needs to reuse information from the initial observation. The model learns the operation-specific LoRA, the \bsc MLP, and the background embedding. The objective is
\begin{equation}
  \mathcal{L}=\mathcal{L}_{\mathrm{flow}}+\lambda\mathcal{L}_{\mathrm{corr}},
\label{eq:loss}
\end{equation}
where $\mathcal{L}_{\mathrm{flow}}$ is the flow-matching loss with increased weight on edited regions. We drop the memory stream with probability $p$ to retain the single-round editing ability. At inference, we apply \cref{eq:round} recursively, using the generated image as the next round's source and updating the persistent memory state throughout the sequence. Our later experiments show that the model trained on two-edit sequences can be directly applied to long editing sequences.

\par\endgroup

\section{Experiments}
\label{sec:experiments}

\begingroup\color{black}

We evaluate \ours on single- and multi-round 3D layout-controlled editing. Single-round comparisons measure spatial control and preservation. Multi-round comparisons and ablations assess how memory and box-local correspondence support consistency.

\subsection{Proposed Dataset}
\label{sec:dataset}

We introduce \data, a multiview dataset for 3D layout-controlled image editing. It provides a training set of two-edit revisit sequences and an evaluation benchmark of single-round edits and ten-round editing sequences. Existing rendered editing datasets provide image pairs for individual 3D edits~\cite{object3dit,boxctrl,tib}, and multi-round datasets cover sequences of object transformations~\cite{ffse} or generated editing trajectories~\cite{mtedit}. In contrast, \data pairs multiview renders with calibrated 3D layouts across object addition, movement, deletion, and camera changes, and includes revisits in scenes with detailed backgrounds.

\noindent\textbf{Scene and state construction.}
Each arrangement places two to five BlenderKit assets in an artist-made BlenderKit scene~\cite{blenderkit}, subject to the available placement area. An arrangement has a base state and up to three edited states: the \emph{added} state inserts an object, the \emph{moved} state translates and rotates an existing object, and the \emph{offset} state moves the same object by 0.2--2\,m on the ground plane and rotates it by at most $45^\circ$. We check placements against rendered scene geometry and flag remaining visible intersections.

\noindent\textbf{Shared cameras and annotations.}
We sample up to 16 views on an arc facing the arrangement, prioritizing views in which the added object and at least one pose of the moved object are visible. The offset pose is chosen so that the object remains visible in most of these views. All states are rendered with Blender EEVEE~\cite{blender} under the scene lighting, and sharing cameras across states yields aligned observations of each edit. Each render includes 3D boxes, camera parameters, and instance masks. Qwen3-VL~\cite{qwen3vl} generates asset categories and captions.

\noindent\textbf{Revisit sequences.}
Transitions between the base and added states yield addition and deletion edits, transitions among the base, moved, and offset states yield object movements, and changes of view yield camera edits. Chaining these transitions produces both exact returns and revisits at an offset pose or a nearby view. In total, \data contains 498 arrangements in 164 scenes, with 247 distinct assets, 6,954 views, and 25,964 renders.

\begin{figure}[t]
  \centering
  \includegraphics[width=\linewidth,trim={6.5bp 5.1bp 0.9bp 8.5bp},clip]{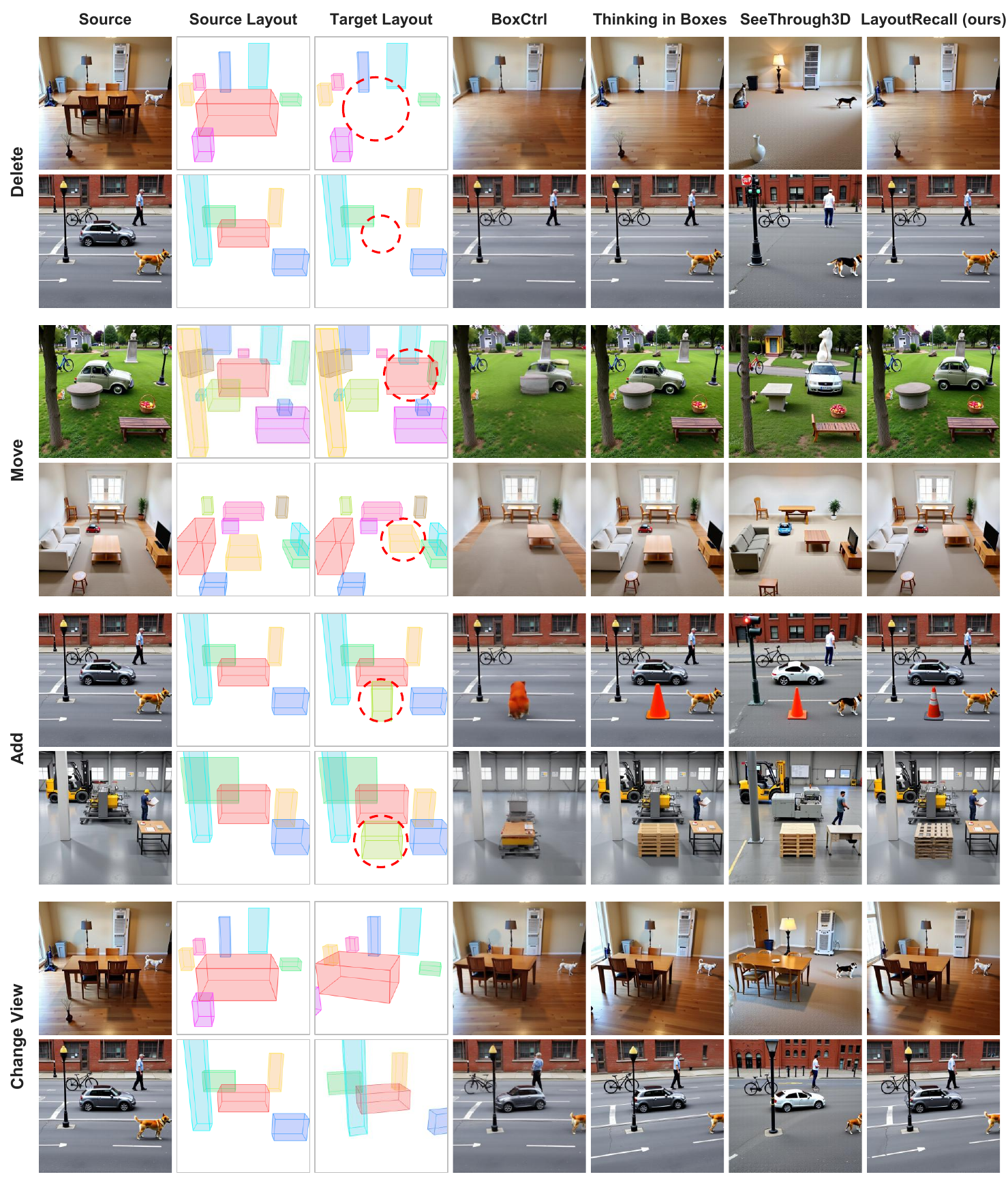}
  \caption{\textbf{Qualitative results of single-round editing.} The prompts for the two added objects are ``an orange traffic cone'' and ``a stack of wooden pallets''.}
  \label{fig:e1}
\end{figure}

\begin{table*}[t]
  \centering
  \small
  \setlength{\tabcolsep}{4pt}
  \caption{\textbf{Multi-round editing on \data.} Back is the PSNR when the camera returns to an earlier view, and Self is the object similarity to its first observation. Best results are in \textbf{bold} and second best are \underline{underlined}.}
  \label{tab:mr}
  \begin{tabular*}{\linewidth}{@{\extracolsep{\fill}}l c cc ccc cc@{}}
    \toprule
    & \multicolumn{1}{c}{Quality} & \multicolumn{2}{c}{Editing Accuracy} & \multicolumn{3}{c}{Preservation} & \multicolumn{2}{c}{Consistency} \\
    \cmidrule(lr){2-2}\cmidrule(lr){3-4}\cmidrule(lr){5-7}\cmidrule(lr){8-9}
    Method & $|\Delta$MUSIQ$|\downarrow$ & mIoU$\uparrow$ & Succ.$\uparrow$ & PSNR$\uparrow$ & SSIM$\uparrow$ & LPIPS$\downarrow$ & Back$\uparrow$ & Self$\uparrow$ \\
    \midrule
    BoxCtrl~\cite{boxctrl} & 18.65 & 0.485 & 0.540 & 17.16 & \underline{0.688} & 0.491 & \underline{15.46} & 0.287 \\
    Thinking in Boxes~\cite{tib} & \underline{4.32} & 0.470 & 0.568 & \underline{18.03} & 0.676 & 0.430 & 14.96 & 0.570 \\
    SeeThrough3D~\cite{seethrough3d} & 7.36 & \textbf{0.618} & \textbf{0.816} & 9.24 & 0.411 & 0.816 & 9.48 & 0.340 \\
    \midrule
    MT-OPSD~\cite{mtopsd} & 10.79 & 0.149 & 0.180 & 15.46 & 0.650 & 0.504 & 12.98 & 0.414 \\
    FreqEdit~\cite{freqedit} & 4.98 & 0.280 & 0.288 & 16.76 & 0.621 & \underline{0.424} & 15.14 & \underline{0.596} \\
    \midrule
    \ours (ours) & \textbf{2.19} & \underline{0.608} & \underline{0.732} & \textbf{24.97} & \textbf{0.802} & \textbf{0.245} & \textbf{23.75} & \textbf{0.733} \\
    \bottomrule
  \end{tabular*}
\end{table*}

\begin{table*}[t]
  \centering
  \small
  \setlength{\tabcolsep}{4pt}
  \caption{\textbf{Multi-round editing on real scenes.} All metrics are reference-free. $|\Delta$MUSIQ$|$ is the mean absolute change in MUSIQ between each output and its input. Survive is the fraction of unedited and visible objects that Grounding DINO still detects inside their boxes after the edit. Best results are in \textbf{bold} and second best are \underline{underlined}.}
  \label{tab:real_mr}
  \begin{tabular*}{\linewidth}{@{\extracolsep{\fill}}l c cc ccc cc@{}}
    \toprule
    & \multicolumn{1}{c}{Quality} & \multicolumn{2}{c}{Editing Accuracy} & \multicolumn{3}{c}{Preservation} & \multicolumn{2}{c}{Consistency} \\
    \cmidrule(lr){2-2}\cmidrule(lr){3-4}\cmidrule(lr){5-7}\cmidrule(lr){8-9}
    Method & $|\Delta$MUSIQ$|\downarrow$ & mIoU$\uparrow$ & Succ.$\uparrow$ & Survive$\uparrow$ & PSNR$\uparrow$ & LPIPS$\downarrow$ & Back$\uparrow$ & Self$\uparrow$ \\
    \midrule
    BoxCtrl~\cite{boxctrl} & 26.05 & 0.442 & 0.530 & 0.191 & \underline{29.86} & 0.113 & \textbf{31.42} & 0.215  \\
    Thinking in Boxes~\cite{tib} & 4.42 & 0.372 & 0.430 & 0.421 & 28.18 & \underline{0.075} & 19.96 & \underline{0.500}  \\
    SeeThrough3D~\cite{seethrough3d} & 7.02 & \underline{0.522} & \underline{0.710} & \underline{0.604} & 12.74 & 0.518 & 15.52 & 0.453  \\
    \midrule
    MT-OPSD~\cite{mtopsd} & 12.74 & 0.118 & 0.130 & 0.104 & 26.66 & 0.090 & 19.80 & 0.394  \\
    FreqEdit~\cite{freqedit} & \underline{4.36} & 0.160 & 0.150 & 0.191 & 26.27 & 0.092 & 25.13 & 0.491  \\
    \midrule
    \ours (ours) & \textbf{2.96} & \textbf{0.662} & \textbf{0.830} & \textbf{0.838} & \textbf{32.84} & \textbf{0.037} & \underline{28.80} & \textbf{0.756}  \\
    \bottomrule
  \end{tabular*}
\end{table*}

\subsection{Experimental Setup}
\label{sec:exp:setup}

\noindent\textbf{Evaluation sequences.}
For synthetic evaluation, we use \data (\cref{sec:dataset}): 400 single-round edits (100 per operation) and 50 ten-round sequences that mix addition, movement, deletion, and camera changes and revisit objects and views after intervening edits. For real-scene evaluation, we select COCO images~\cite{coco} in which every major object has a matching WildDet3D 3D box annotation~\cite{wilddet3d}, yielding 80 single-round edits (20 per operation) and 10 ten-round sequences.

\noindent\textbf{Compared methods.}
We compare with SeeThrough3D~\cite{seethrough3d}, a 3D layout-conditioned generator, and with BoxCtrl~\cite{boxctrl} and Thinking-in-Boxes~\cite{tib}, two single-round 3D layout-controlled editors, in both single-round and multi-round editing. All three receive the same underlying geometry in their respective layout formats; SeeThrough3D generates from the target layout alone, whereas the two editors also condition on the current image. For multi-round editing, we additionally compare with MT-OPSD~\cite{mtopsd} and FreqEdit~\cite{freqedit}, two general multi-turn image editors, which receive text instructions derived from the same edits. In multi-round editing, every method runs recursively on its own outputs.

\begin{figure*}[t]
  \centering
  \includegraphics[width=\linewidth,trim={0bp 20cm 2cm 5bp},clip]{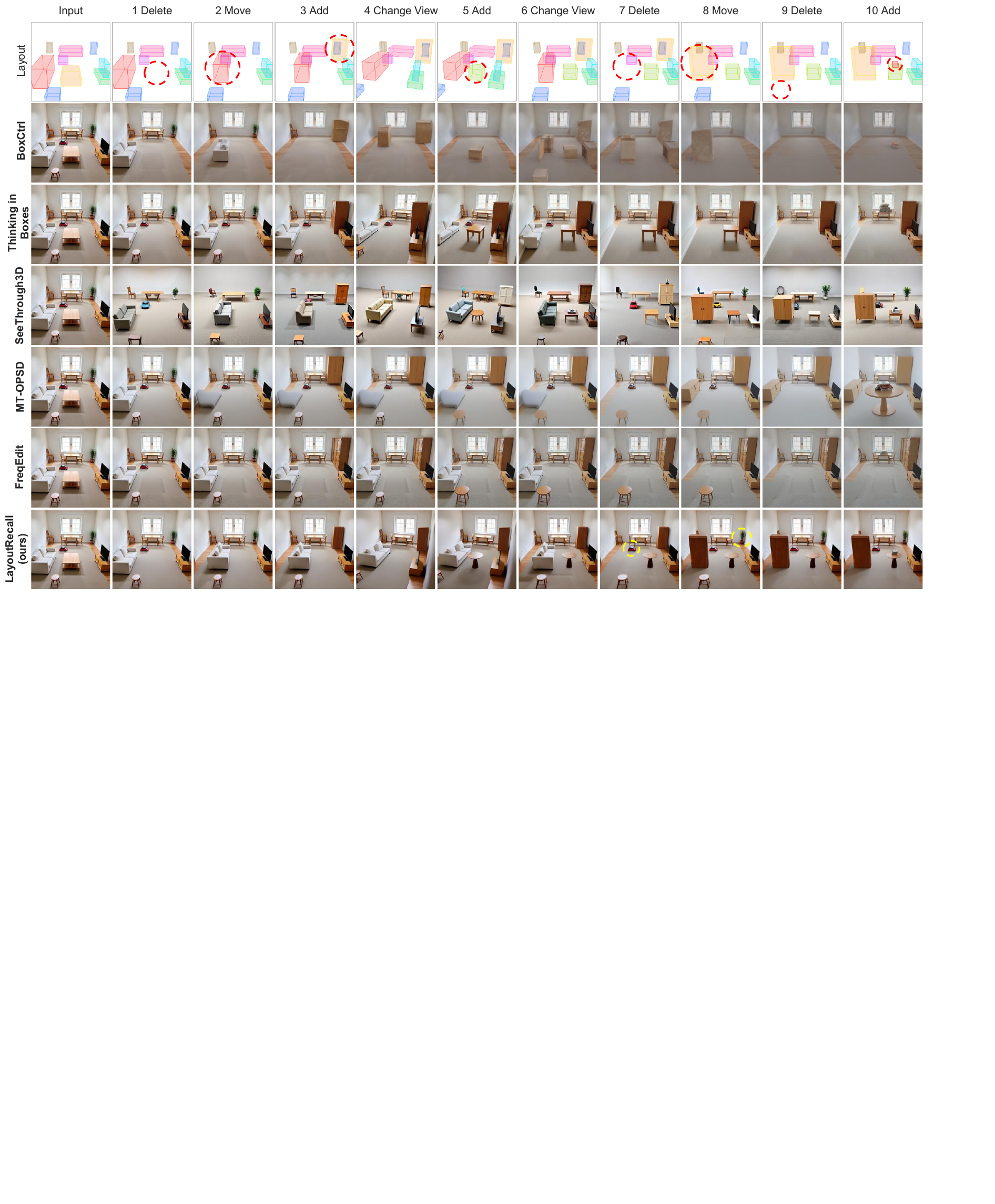}
  \caption{\textbf{Qualitative results of multi-round editing.} SeeThrough3D regenerates the target while completely discarding the original content. BoxCtrl replaces the scene with an empty background. With its long-term memory, \ours restores the toy car previously hidden by the moved sofa in round~7 and the plant previously hidden by the added cabinet in round~8 (yellow circles).}
  \label{fig:mr}
\end{figure*}

\noindent\textbf{Metrics.}
On \data, image quality is measured by $|\Delta$MUSIQ$|$, the mean absolute difference between the MUSIQ scores~\cite{musiq} of each output and its reference render. Single-round editing accuracy is measured per operation: CLIP-T~\cite{clip} for addition, DINOv2~\cite{dinov2} similarity between source and target crops for movement, DINOv2 similarity between the output and the reference within the projected box of the removed object for deletion, and full-image DINOv2 similarity to the reference for camera changes. Preservation is measured by PSNR, SSIM, and LPIPS~\cite{lpips} outside the edited regions. Grounding DINO~\cite{groundingdino} detections provide the mIoU with the projected target boxes and the edit success rate. For consistency, Back computes PSNR when the camera returns to an earlier view, and Self computes the DINOv2 similarity between an object and the model's first observation of the same object.
Since real scenes have no edited references, we use only reference-free metrics on them. Preservation compares each output with the method's own source image outside the boxes of the edited object, and Back compares a camera return with the method's own earlier output at that view, outside all boxes. Survive is the fraction of unedited and visible objects that Grounding DINO still detects inside their boxes after the edit. $|\Delta$MUSIQ$|$ is measured against the input photograph instead of a reference render; lower values indicate that the editor better retains the photographic quality of the input.

\noindent\textbf{Implementation.}
We build on FLUX.2-klein-9B-base with rank-16 LoRA ($\alpha=32$). Each operation-specific adapter is trained on 20k sequences with a learning rate of $10^{-4}$ and 10 gradient accumulation steps. The retrieved memory is capped at 2048 tokens per round. We set the object assignment threshold to $\tau_{\mathrm{obj}}=20\%$, the matching radius to $\rho=1.5$ token footprints, the correspondence loss weight to $\lambda=0.025$, and the memory dropout probability to $p=0.15$. The training of a LoRA checkpoint takes 15 hours on an NVIDIA RTX PRO 6000 Blackwell GPU, and an editing round at inference takes 24 seconds.

\subsection{Single-Round Editing}
\label{sec:exp:e1}

\Cref{tab:e1,tab:real_sr} report single-round results on synthetic and real scenes, respectively. On real scenes, \ours achieves the best scores on all metrics. On \data, it achieves the best image quality, the best preservation, and the highest movement, deletion, and camera accuracy. SeeThrough3D scores higher on addition, but as a pure generation model it discards the input image and synthesizes a new one, and therefore obtains by far the lowest preservation scores. Qualitative results are shown in \cref{fig:e1}. Overall, adding the memory module does not compromise single-round editing: \ours performs on par with or better than methods designed specifically for this setting. We attribute this to dataset fine-tuning and memory dropout.

\subsection{Multi-Round Editing}
\label{sec:exp:e2}
\label{sec:exp:e3}

\vspace{-2mm}

\Cref{tab:mr,tab:real_mr} report multi-round results. \ours achieves the best results on most metrics. On \data, SeeThrough3D obtains higher editing accuracy at the cost of very low preservation. On real scenes, BoxCtrl replaces each scene with an empty background in the first round, which inflates its PSNR and Back scores, but leads to by far the worst $|\Delta$MUSIQ$|$. Without access to 3D layout information, the general multi-turn editors (MT-OPSD and FreqEdit) rarely follow the editing instructions or produce the intended scene, resulting in very low editing accuracy. The single-round 3D layout-controlled methods obtain low preservation and consistency scores, since they retain no information about objects from previous rounds. In contrast, \ours executes the editing instructions precisely and maintains consistency throughout the editing process. Moreover, because the appearance stored in memory is injected at every round, \ours is less susceptible to image quality degradation and achieves a high overall image quality. \Cref{fig:mr} shows consistent appearance when objects reappear (yellow circles).

\subsection{Ablation Study}
\label{sec:exp:ablation}

\begin{table}[t]
  \centering
  \small
  \setlength{\tabcolsep}{3pt}
  \caption{\textbf{Ablations on \data.} Best results are in \textbf{bold} and second best are \underline{underlined}.}
  \label{tab:e4}
  \resizebox{\linewidth}{!}{%
  \begin{tabular}{l c c cc}
    \toprule
    & \multicolumn{1}{c}{Quality} & \multicolumn{1}{c}{Preservation} & \multicolumn{2}{c}{Consistency} \\
    \cmidrule(lr){2-2}\cmidrule(lr){3-3}\cmidrule(lr){4-5}
    Variant & $|\Delta$MUSIQ$|\downarrow$ & PSNR$\uparrow$ & Back$\uparrow$ & Self$\uparrow$ \\
    \midrule
    ours & \textbf{2.19} & \textbf{24.97} & \textbf{23.75} & \textbf{0.733} \\
    w/o \bsc & \textbf{2.19} & \underline{24.24} & \underline{22.54} & \underline{0.721} \\
    w/ last 2 frames & 2.70 & 22.96 & 19.18 & 0.678 \\
    w/o mem & 3.73 & 20.18 & 15.71 & 0.601 \\
    \bottomrule
  \end{tabular}}
\end{table}

\begin{table}[t]
  \centering
  \small
  \setlength{\tabcolsep}{3pt}
  \caption{\textbf{Ablations on real scenes.} All metrics are reference-free. Best results are in \textbf{bold} and second best are \underline{underlined}.}
  \label{tab:real_e4}
  \resizebox{\linewidth}{!}{%
  \begin{tabular}{l c c cc}
    \toprule
    & \multicolumn{1}{c}{Quality} & \multicolumn{1}{c}{Preservation} & \multicolumn{2}{c}{Consistency} \\
    \cmidrule(lr){2-2}\cmidrule(lr){3-3}\cmidrule(lr){4-5}
    Variant & $|\Delta$MUSIQ$|\downarrow$ & PSNR$\uparrow$ & Back$\uparrow$ & Self$\uparrow$ \\
    \midrule
    ours & \underline{2.96} & \textbf{32.84} & \textbf{28.80} & \textbf{0.756}  \\
    w/o \bsc & \textbf{2.28} & \underline{32.58} & \underline{28.60} & \underline{0.753}  \\
    w/ last 2 frames & \underline{2.96} & 32.42 & 24.57 & 0.648  \\
    w/o mem & 4.02 & 32.23 & 16.55 & 0.596  \\
    \bottomrule
  \end{tabular}}
\end{table}

\begin{figure*}[t]
  \centering
  \includegraphics[width=\linewidth,trim={0bp 0bp 1.5cm 0bp},clip]{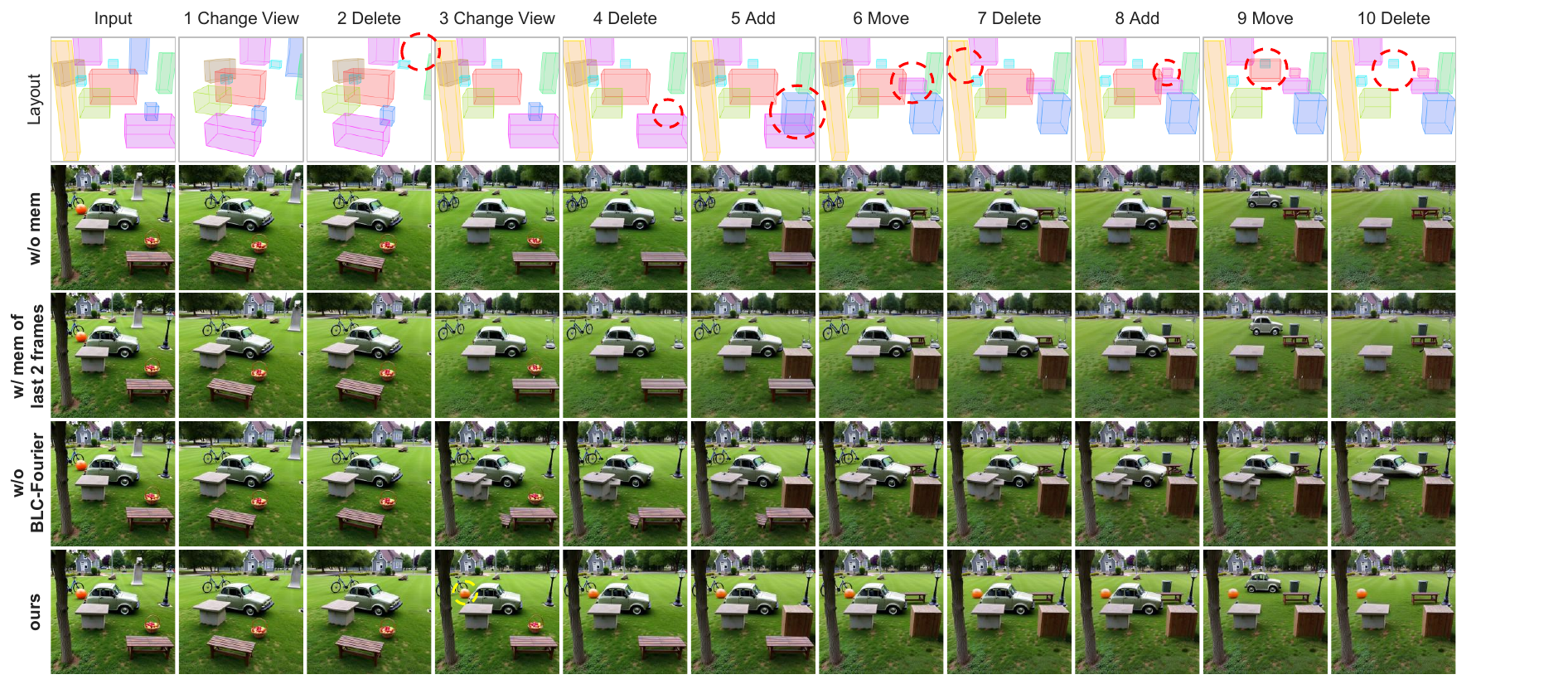}
  \caption{\textbf{Qualitative ablation results.} When the camera returns in round~3, only the full model restores the orange ball (yellow circle). Without \bsc, the stone table and the bike are distorted.}
  \label{fig:ablation}
\end{figure*}

We compare the full model with three variants: without \bsc, with memory from only the last two frames, and without memory. Results are shown in \cref{tab:e4,tab:real_e4,fig:ablation}.

Removing the memory causes substantial drops in image quality, preservation, and consistency. Keeping memory from only the last two frames yields scores between those of the memory-free variant and the full model, as expected. These results confirm the importance of long-term memory for maintaining consistency over long editing sequences. Removing \bsc moderately reduces most scores and can distort objects (\cref{fig:ablation}), supporting its role in establishing spatial correspondence.

\par\endgroup

\section{Conclusion}
\label{sec:conclusion}
\begingroup\color{black}

\ours extends 3D layout-controlled image editing to sequences that revisit earlier objects and views. \bslm stores historical appearance in box-local surface coordinates, and \gamc conditions the editing on observations projected into the requested scene with spatial and semantic alignments. \caloss encourages the editor to use the matching surface evidence. Through extensive experiments, we show that our model preserves object appearance more effectively than other models and ablation variants with shorter or empty memory, while retaining the ability to produce high-quality outputs and follow spatial edits precisely.

\vspace{1mm}

\noindent\textbf{Limitations.}
The box surface approximates object geometry, so neighboring-cell retrieval can introduce appearance in empty parts of a loose box. Also, background recovery depends on previously observed camera views.
\par\endgroup

\section*{LLM Usage Statement}
\label{sec:llm}

We used LLM-based assistants (Claude via Claude Code) to help implement and debug code, run and analyze experiments, prepare figures, and draft and revise the manuscript. All research ideas, design decisions, and reported results were determined by the authors, who reviewed all assistant outputs and take full responsibility for the content of this paper.

{
    \small
    \bibliographystyle{ieeenat_fullname}
    \bibliography{main}
}

\clearpage
\maketitlesupplementary

\appendix
\renewcommand{\thesection}{\Alph{section}}

This supplementary material provides the implementation details of \ours (\cref{supp:impl}), the full evaluation protocol (\cref{supp:metrics}), additional qualitative results of multi-round editing (\cref{supp:qualitative}), an analysis of image degradation over long editing sequences (\cref{supp:long}), and the details of our dataset \data (\cref{supp:dataset}).

\section{Implementation Details}
\label{supp:impl}

\noindent\textbf{Backbone and inputs.}
\ours builds on FLUX.2-klein-9B-base~\cite{flux2}, a flow-matching transformer with 8 double-stream blocks, 24 single-stream blocks, and a hidden size of 4096. The FLUX.2 VAE encodes each $16\times16$ image patch into one token with 128 channels. The text prompt is the short caption of the new object for addition and a fixed prompt for the other operations. In each round, the transformer jointly attends to the text tokens, the noisy target tokens, the reference streams $B_{t-1}$, $B_t$, and $I_{t-1}$ (with RoPE time offsets of 10, 20, and 30), and the retrieved memory tokens (offset 40).

\noindent\textbf{Operation-specific models.}
Each operation has its own LoRA adapter (rank 16, $\alpha=32$) on the attention projections and MLPs of all transformer blocks, together with its own \bsc MLP and background embedding. Since addition introduces a new object without any stored appearance, it uses the single-round adapter without memory, whereas deletion, movement, and camera changes use the memory adapters. \Cref{tab:hparams} summarizes the hyperparameters.

\begin{table}[t]
  \centering
  \small
  \setlength{\tabcolsep}{4pt}
  \caption{\textbf{Hyperparameters of \ours.}}
  \label{tab:hparams}
  \begin{tabular}{@{}p{0.45\linewidth}p{0.51\linewidth}@{}}
    \toprule
    \multicolumn{2}{@{}l}{\emph{Box-Surface Latent Memory}} \\
    Cell size & 8\,cm on each box face \\
    Object token threshold $\tau_\mathrm{obj}$ & 20\% mask coverage \\
    Replacement margin & $1.25\times$ projected area \\
    Memory budget & 1024 (training), 2048 (inference) \\
    \midrule
    \multicolumn{2}{@{}l}{\emph{Instance masks (SAM~3~\cite{sam3})}} \\
    Text prompts & category and short caption \\
    Accepted instance & score $\geq0.3$, $\geq60\%$ inside box \\
    Weak instance & score $\geq0.02$, $\geq90\%$ inside box \\
    Hidden object & $>50\%$ of box behind other boxes \\
    \midrule
    \multicolumn{2}{@{}l}{\emph{Geometry-Aligned Memory Conditioning}} \\
    \bsc & 39 $\to$ 1024 $\to$ 4096, zero-initialized output \\
    Object attention region & box coverage $\geq5\%$, dilated by 1 token \\
    Background attention region & box coverage $<50\%$, dilated by 1 token \\
    \midrule
    \multicolumn{2}{@{}l}{\emph{Training}} \\
    Steps & 20k sequences \\
    Batch & 1, 10 gradient accumulation steps \\
    Optimizer & AdamW, lr $10^{-4}$, weight decay 0.01 \\
    Schedule & cosine, 50 warm-up updates \\
    Timesteps & $t\sim\mathcal{U}[0,1]$ \\
    Loss weight of edited region & 5 \\
    Memory dropout $p$ & 0.15 \\
    \caloss & $\lambda=0.025$, $\beta=0.3$, $\rho=1.5$ tokens, \\
     & 8 double blocks $\times$ first 8 heads \\
    \midrule
    \multicolumn{2}{@{}l}{\emph{Inference}} \\
    Sampling & 50 steps, no classifier-free guidance \\
    \bottomrule
  \end{tabular}
\end{table}

\noindent\textbf{Memory writing.}
SAM~3 receives the category and caption of each object as text prompts. An instance is accepted as the object mask if its score is at least 0.3 and at least 60\% of its pixels lie inside the projected box; a weaker instance with a score of at least 0.02 is accepted if at least 90\% of its pixels lie inside the box. An object more than half covered by the boxes of other objects is treated as hidden and is not written. Masks are clipped to the projected boxes, and a pixel claimed by several objects is assigned to the box that is nearest along its ray. For each token assigned to an object, we intersect the camera ray through the token center with the box of that object. If several tokens hit the same cell, the token nearest to the cell center is written. A stored cell is replaced only by an observation whose projected area is at least $1.25\times$ larger, \ie, an observation at a higher image resolution. Deleting an object also removes its memory. Background tokens not covered by any mask or box are stored once per camera view and image location.

\noindent\textbf{Memory retrieval.}
The ray of each target token is intersected with the front-most target box. If the hit cell is stored, it is used directly; otherwise, with a lookup radius of 1, we use the nearest stored cell within one cell on the same face. Each retrieved cell contributes one memory token, positioned at the projection of its cell center under the target box and camera. Background memory is retrieved from the target camera view if that view has been observed. If the target view is that of $I_{t-1}$, we retrieve the stored background tokens that were covered by the boxes in $I_{t-1}$ and are visible in the target. When the retrieved tokens exceed the memory budget, high-frequency tokens are kept first.

\noindent\textbf{Warped background.}
For an unobserved camera view, the camera model warps the stored views into the target view, using a ground plane at the lowest box bottom and a backdrop sphere around the scene. A warped token is kept if at least 90\% of its pixels are valid and its reprojection error, estimated by varying the sphere radius and the ground height, is at most one token. The warped composite is encoded by the VAE and receives a separate learned background embedding.

\noindent\textbf{Training.}
Each LoRA checkpoint is trained on 20k two-edit sequences $(I_0,I_1,I_2)$. The ground-truth frames $I_0$ and $I_1$ are written into memory, and the loss is computed only on the second edit. The memory stream is dropped with probability $p=0.15$. The base model, VAE, text encoder, and SAM~3 remain frozen. Training one memory adapter takes 15 hours on a single NVIDIA RTX PRO 6000 Blackwell GPU.

\noindent\textbf{Inference.}
Inference is recursive: each round takes the previous output of the model as $I_{t-1}$, the masks of generated frames are computed online with SAM~3, and each generated frame is written into memory at the end of its round. One editing round takes 24 seconds.

\section{Evaluation Protocol}
\label{supp:metrics}

\subsection{Synthetic Scenes}

\noindent\textbf{Projected regions.}
We project the eight corners of a box through the camera and clip their bounding rectangle to the image. For object-level metrics, this crop is padded to a square and resized to $224\times224$. When the camera is fixed, we identify the edited pixels from the RGB difference between the source and target reference renders and dilate the resulting mask by 8 pixels:
\begin{equation}
 M_{\mathrm{edit}}=\mathrm{dilate}\!\left(\left|I^{\mathrm{ref}}_{\mathrm{src}}-I^{\mathrm{ref}}_{\mathrm{tgt}}\right|>0.04\right),
\end{equation}
where a change in any channel suffices. The mask thus also covers secondary effects such as shadows and reflections, and preservation (PSNR, SSIM, and LPIPS~\cite{lpips}) is measured on its complement. In single-round editing, preservation is computed for addition, deletion, and movement; camera changes are excluded because they alter the whole image.

\noindent\textbf{Per-operation editing accuracy.}
\Cref{tab:supp_ops} defines the per-operation editing accuracy metrics.

\begin{table}[t]
  \centering
  \small
  \caption{\textbf{Per-operation editing accuracy on \data.}}
  \label{tab:supp_ops}
  \begin{tabular}{@{}lp{0.74\linewidth}@{}}
    \toprule
    Operation & Metric \\
    \midrule
    Addition & CLIP-T~\cite{clip}: crop at the target box vs.\ ``a photo of \{caption\}'' \\
    Movement & DINOv2~\cite{dinov2} cosine similarity: crop at the target box vs.\ crop at the source box of the source image \\
    Deletion & DINOv2 cosine similarity: output vs.\ reference render, both cropped at the box of the removed object \\
    Camera & DINOv2 cosine similarity: full output vs.\ full reference render \\
    \bottomrule
  \end{tabular}
\end{table}

\noindent\textbf{Detection-based editing accuracy.}
We query Grounding DINO~\cite{groundingdino} with the object category, using box and text thresholds of 0.3 and 0.25. For addition and movement, we take the maximum IoU between any detection and the projected target box. For a camera change, we average these per-object maxima over all visible objects. The mIoU is averaged over addition, movement, and camera changes, and each of these edits counts as a success when its IoU exceeds 0.5. A deletion succeeds when every detection of the deleted category has an IoU of at most 0.3 with the source box.

\noindent\textbf{Consistency.}
An object is considered visible when at least half of its pixels are visible according to the reference instance masks. Back is the PSNR at rounds that return to an earlier camera view, computed against the reference render before the camera change and excluding the projected boxes of objects edited in between. Self is the DINOv2 cosine similarity between the crop of each visible object and its crop at its first visible observation in the outputs of the same method, averaged over objects and rounds.

\noindent\textbf{Image quality.}
We compute MUSIQ~\cite{musiq} (trained on KonIQ-10k) for every output and its reference render at the evaluation resolution. $|\Delta$MUSIQ$|$ is the absolute difference between the two scores, averaged over all outputs of a method. Unlike the raw score, the absolute difference penalizes deviations in either direction: degradation lowers MUSIQ, whereas global brightening or regenerating the scene can raise it without preserving the input.

\subsection{Real Scenes}

\noindent\textbf{Evaluation data.}
We start from the COCO~\cite{coco} images annotated with 3D boxes in WildDet3D~\cite{wilddet3d}. An image is a candidate if it has at most 10 valid 3D boxes and every COCO object covering at least 1\% of the image is matched by a 3D box whose projection has an IoU of at least 0.5 with the COCO box. Among the candidates with 3--10 boxes inside the central square crop, we select 20 images by visual inspection, requiring a visible ground plane, tight boxes, and a box for every major object, and we remove duplicate boxes. Each image is center-cropped to a square. The boxes are converted to gravity-aligned boxes on a common ground plane whose up direction is the mean vertical axis of the boxes, and each scene is uniformly rescaled to typical object heights, which leaves all projections unchanged. The converted boxes reproject within a median of 5 pixels of the original annotations. Edits are generated automatically: deleting a visible object, moving a standing object by 0.8--1.8 times its size, adding one of eight everyday objects next to an existing object, and orbiting the camera by 10--18$^\circ$. Every edit is checked to stay in view and to be free of collisions.

\noindent\textbf{Reference-free metrics.}
Since real scenes have no reference renders of the edited states, we use reference-free metrics. Preservation compares each output with the source image of the same method, excluding the boxes of the edited object in both frames, dilated by 8 pixels. Back compares an output at a revisited camera view with the earlier output of the same method at that view, excluding all boxes in both frames. Survive is the fraction of unedited, visible objects that Grounding DINO still detects inside their target boxes (IoU above 0.5) after the edit. Detection-based accuracy and Self follow the synthetic protocol, and $|\Delta$MUSIQ$|$ is measured against the input photograph.

\section{Additional Qualitative Results}
\label{supp:qualitative}

\begin{figure*}[t]
  \centering
  \includegraphics[width=\linewidth,trim={0 0 2cm 0},clip]{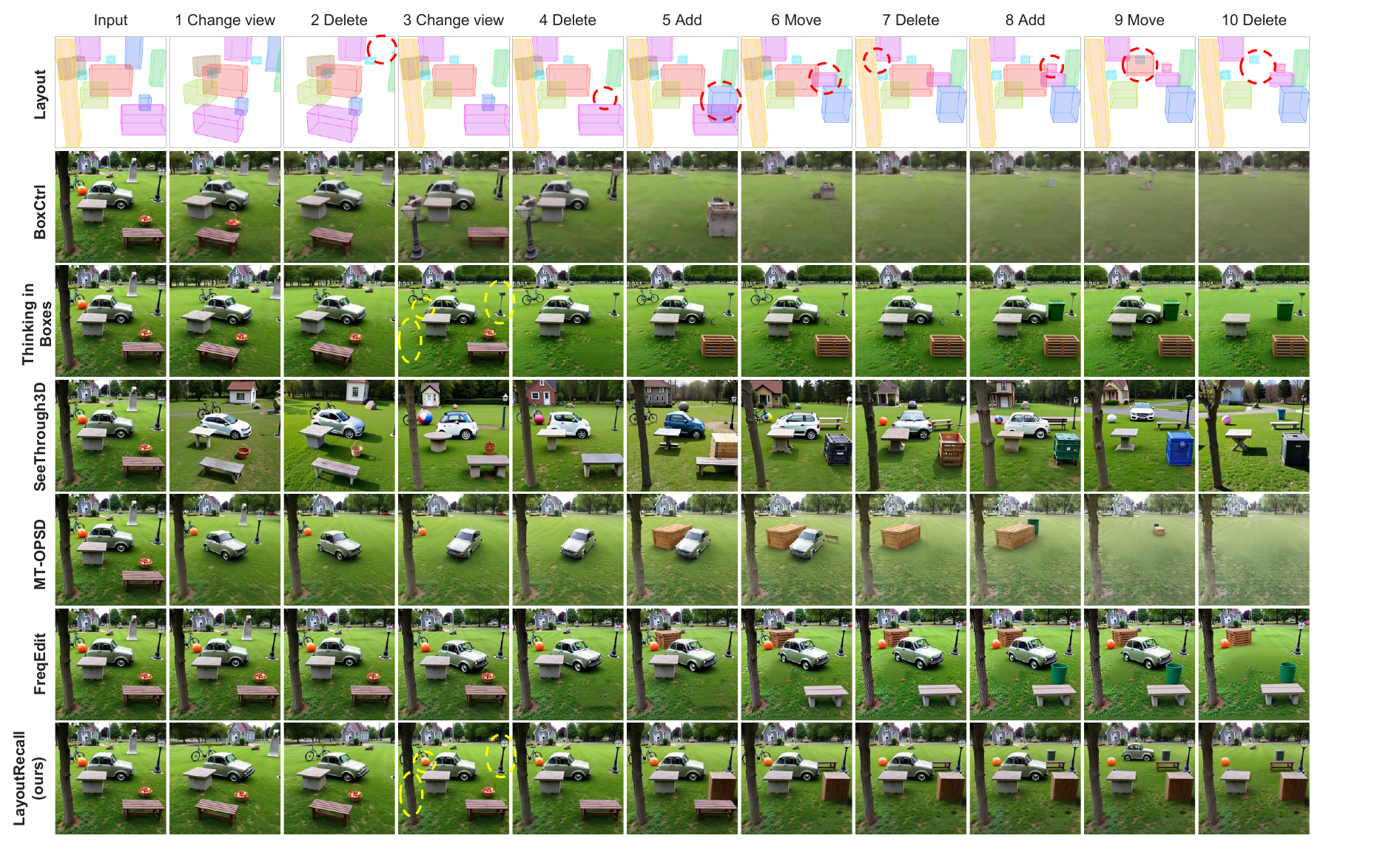}
  \caption{\textbf{Additional qualitative results of multi-round editing.} In round~3 (yellow circles), unlike the single-round editor Thinking-in-Boxes~\cite{tib}, \ours restores the appearance of the orange ball, which was occluded by the car, and of the streetlight and the tree, which were out of view.}
  \label{fig:supp:mr}
\end{figure*}

\Cref{fig:supp:mr} shows an additional multi-round editing sequence. The camera orbit in round~1 hides the orange ball behind the car and moves the tree and the streetlight out of view, and moving the car back in round~9 hides the rock. \ours restores the ball, the tree, and the streetlight when the camera returns in round~3, and the rock when the car is deleted in round~10, while the background and the remaining objects stay consistent throughout the sequence.

\section{Image Degradation over Long Editing Sequences}
\label{supp:long}

A major challenge in multi-round image editing is error accumulation. Because each round operates on the output of the previous one, small artifacts accumulate over a long editing sequence and eventually degrade the image. We examine this effect for different methods.

\begin{figure}[t]
  \centering
  \includegraphics[width=\linewidth, trim={0 2cm 2cm 0}, clip]{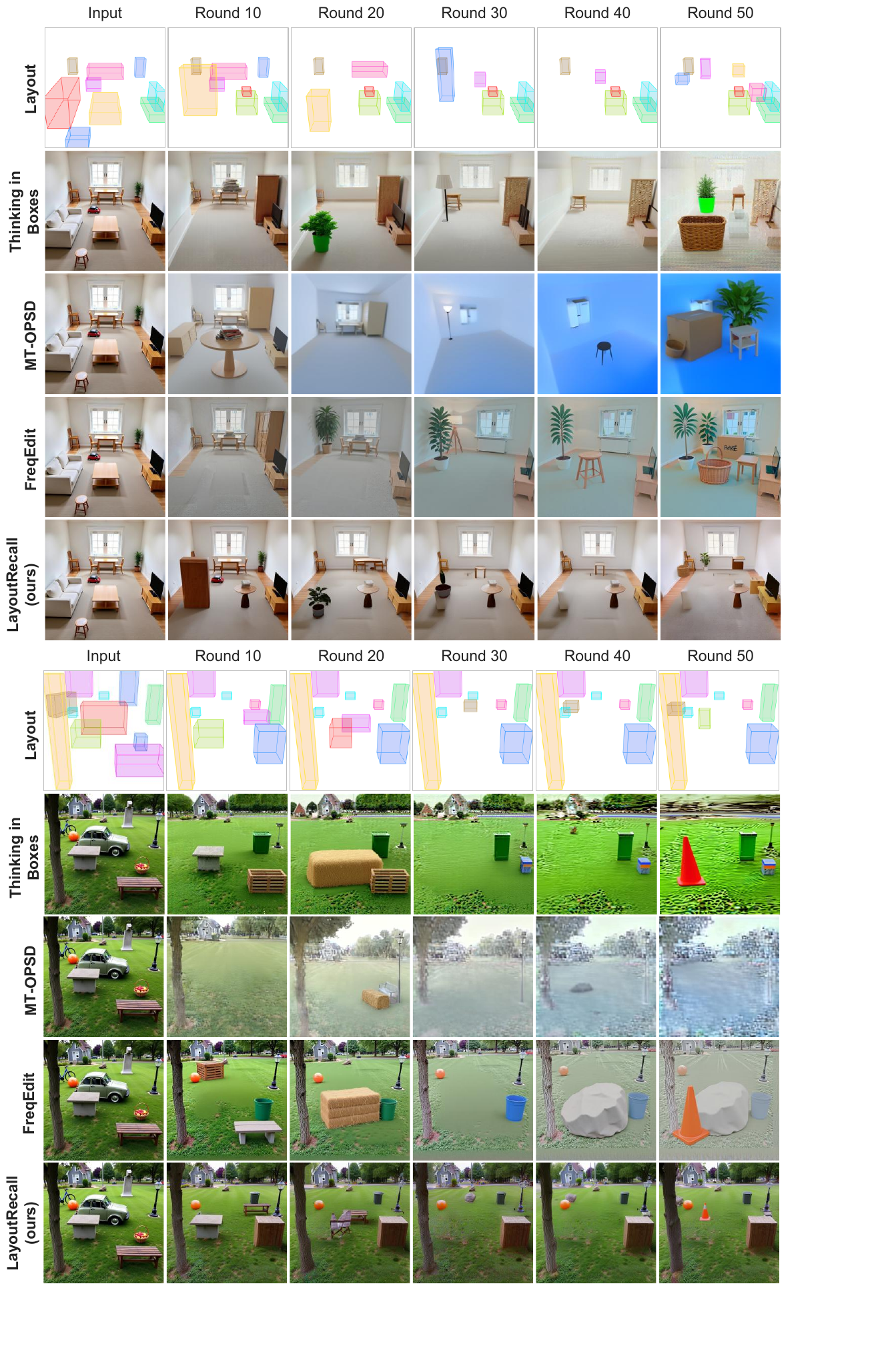}
  \caption{\textbf{Image degradation over long editing sequences.} Outputs of each method over 50 editing rounds. All methods accumulate errors, while \ours stays the cleanest.}
  \label{fig:supp:long}
\end{figure}

\noindent\textbf{Results.}
We run \ours, Thinking-in-Boxes~\cite{tib}, MT-OPSD~\cite{mtopsd}, and FreqEdit~\cite{freqedit} for 50 rounds and show the results in \cref{fig:supp:long}. All methods accumulate errors and degrade as the number of rounds increases, but the outputs of \ours remain the cleanest.

\noindent\textbf{Memory as a high-quality anchor.}
We attribute this robustness to the memory design. Without memory, an editor conditions only on its previous output, so an artifact, once it appears, is propagated and amplified in every later round. In \ours, the tokens retrieved from \bslm enter the denoising process of every round as clean references for the visible objects and the background, so the generation is guided by the stored appearance in addition to the possibly degraded previous output. Moreover, \bslm keeps only high-resolution observations: a stored token is replaced only by an observation of the same surface cell at a substantially higher image resolution (\cref{tab:hparams}), and background tokens are written once per view and image location. Later observations at the same or a coarser resolution, including degraded ones, therefore never overwrite the stored appearance, which typically comes from the input image or the sharpest earlier view. The memory tokens thus act as a high-quality anchor that pulls each round back toward the established appearance and alleviates error accumulation over long sequences.

\section{Dataset Details}
\label{supp:dataset}

\begin{figure*}[t]
  \centering
  \includegraphics[width=\linewidth]{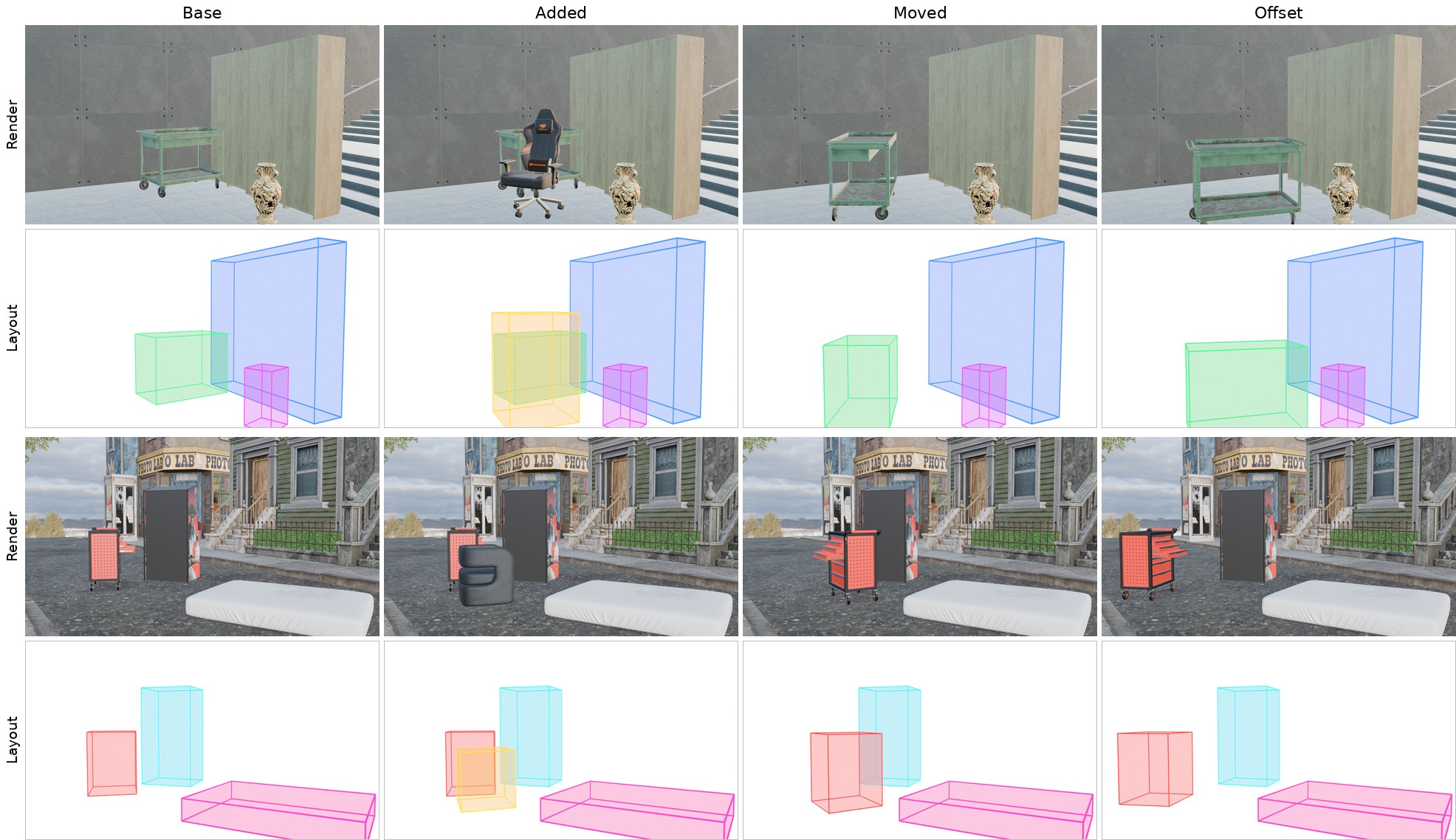}
  \caption{\textbf{Two arrangements of \data.} Each column shows one state from the same camera, with its layout image below. In both arrangements, the added object hides part of the tool cart, so deleting it reveals the cart again.}
  \label{fig:supp:dataset}
\end{figure*}

\Cref{tab:dataset} lists the statistics of \data; this section gives its rules and parameters. \Cref{fig:supp:dataset} shows two arrangements.

\begin{table}[h]
  \centering\color{black}
  \small
  \setlength{\tabcolsep}{4pt}
  \captionsetup{labelfont={color=black}}
  \caption{\color{black}\textbf{Statistics of \data.} Values in parentheses give the 10th--90th percentile range.}
  \label{tab:dataset}
  \begin{tabular}{@{}p{0.40\linewidth}p{0.56\linewidth}@{}}
    \toprule
    \multicolumn{2}{@{}l}{\emph{Scale}} \\
    Scenes, arrangements & 164, 498 \\
    Assets used & 247 (208 in base layouts, 39 only added) \\
    Views & 6,954 \\
    Renders & 25,964 (base, moved and added 6,954 each; offset 5,102) \\
    Scene type & interior 78, exterior 86 \\
    Asset categories & 126 \\
    \midrule
    \multicolumn{2}{@{}l}{\emph{Arrangements}} \\
    Arrangements per scene & 1: 15 scenes, 2: 48, 3: 17, 4: 84 \\
    Objects in the base state & 2: 178, 3: 290, 4: 17, 5: 13 \\
    Views per arrangement & 16 in 337, 3--15 in 161 \\
    Size of the added object & tiny 27, small 342, medium 125, large 4 \\
    \midrule
    \multicolumn{2}{@{}l}{\emph{Edits from the base pose}} \\
    Moved, distance & median 1.39\,m (0.59--3.08\,m) \\
    Moved, $|\Delta\psi|$ & median $89^\circ$ ($12^\circ$--$162^\circ$) \\
    Offset, distance & 0.2--2\,m, 40--42 arrangements per 0.2\,m bin \\
    Offset, $|\Delta\psi|$ & median $21^\circ$, maximum $44.9^\circ$ \\
    \midrule
    \multicolumn{2}{@{}l}{\emph{Cameras}} \\
    Azimuth span per arrangement & median $63^\circ$ ($17^\circ$--$175^\circ$) \\
    Elevation & median $14^\circ$ ($7^\circ$--$25^\circ$) \\
    Distance to the objects & median 4.7\,m (3.3--6.5\,m) \\
    Camera model (scenes) & perspective 155, orthographic 6, panoramic 3 \\
    \midrule
    \multicolumn{2}{@{}l}{\emph{Visibility (474 arrangements)}} \\
   Object area in the image & median 3.5\% (0.5--11.9\%) \\
    Added object hides another & $\geq$10\% in 805 of 6,613 views, $\geq$50\% in 255 \\
    Movement changes another & $\geq$10\% in 1,153 views (offset 658 of 4,911) \\
    Moved object hidden or revealed & 763 views (offset 461) \\
    \bottomrule
  \end{tabular}
\end{table}

\noindent\textbf{Scenes and assets.}
We process 261 BlenderKit scenes.
Of these, 165 yield object arrangements and 164 have rendered views.
The others have no placeable region (33), no arrangement that places or shows a visible change (19), no camera (8) or no readable lighting (8), or exceed the rendering budget (28).
The asset pool holds 1,000 BlenderKit models in four size bins: 209 tiny (longest side below 0.35\,m), 352 small (footprint side below 1\,m, height below 1.5\,m), 251 medium (below 2.5\,m and 3\,m) and 188 large.
We cast rays through a grid from the scene camera and keep hits on upward-facing surfaces; a cluster of these points is the placeable region, and $w$ is its larger horizontal side.
We draw $m=\mathrm{clip}(\lfloor 0.16\,w^2/\bar{A}\rfloor-1,\,2,\,5)$ assets for a scene, where $\bar{A}$ is the mean footprint area of eight provisionally drawn assets, and one more for the added state; the 21 earliest scenes use three.
A scene holds up to four arrangements, and an arrangement can use fewer of these assets when placement fails.
Qwen3-VL-8B~\cite{qwen3vl} generates the category and captions of each asset from its turntable render (missing for 12 of the 247 assets used), its BlenderKit thumbnail, and its BlenderKit category and tags.

\noindent\textbf{Placement.}
We place objects from the longest footprint side to the shortest, at random points of the region and with a yaw drawn uniformly from $[-180^\circ,180^\circ]$.
A first-pass pose is accepted when camera-ray samples of the region support its footprint, and its footprint, padded by 3.5\,cm, and its axis-aligned box overlap those of the other objects by at most 1.5\,cm.
The moved state re-draws the pose of one object at least $\max(\min(0.6\,d,\,0.25\,w),\,0.15\,\text{m})$ from its base position, where $d$ is the longer footprint side.
The added state places one further object against the base layout.
We then audit the base, moved and added states of every arrangement for visible interpenetration with rigid scene geometry, which flags 254 of the 498 arrangements at 1\,cm or more.
We regenerate flagged arrangements where possible: 206 are regenerated, 48 remain flagged, and the other 244 keep their first-pass layout.
A regenerated pose must also have rendered support under its footprint, with a gap of at most 0.3\,m, and pass a collision test.
This test uses the scene geometry as rendered, including instances and render-time displacement.
It rejects crossings of rigid geometry more than 1\,cm above the bottom of the object (up to 15\,cm for small instanced ground cover), containment in either direction and overlap with placed objects; vegetation crossings are recorded but not rejected.

\noindent\textbf{Views.}
Each scene has a budget of 16 views (8 or 12 in 34 scenes).
We generate 1.5 candidate cameras per view on a nominal $97^\circ$ arc centered on the direction of the scene camera, aimed at the union of the base, moved and added boxes, at a distance that fits this union in the frame.
A candidate without a clear, well-framed view moves to the nearest azimuth that has one, so views can span more than the arc.
We shortlist the candidates that best see the moved and added objects, four more than the budget.
An object scores the fraction of nine rays, to its box center and corners, that scene geometry does not block, scaled by $\min(1,\,\text{area}/0.4\%)$, where area is the share of the image its box covers; the moved object takes the better of its two poses.
We keep candidates in which both objects score at least 0.34, or 0.17 if too few pass, and spread up to the budget of views over them.
The 51 arrangements of the 19 earliest scenes predate this selection.
After rendering, we drop a view in which fewer than 200 pixels change by more than 10/255 between the base state and the moved or added state.
Perspective views use a 35\,mm lens on a 36\,mm sensor, except for one scene keyframed to 24\,mm and two with other sensors, and 13 scenes keep a lens shift.
Six scenes keep an orthographic camera, and three keep a panoramic setting that EEVEE renders as pinhole.

\noindent\textbf{Offset state.}
To fill the distance bins evenly, a feasibility search tests every arrangement in each 0.2\,m bin, with directions every $15^\circ$ and radii across the bin.
We assign arrangements to bins, most constrained first, into the bin with the largest deficit, and draw the turn uniformly from $[-45^\circ,45^\circ]$.
An offset pose must pass the collision test above, leave at most 5\,cm between object and support, change height by at most 0.45\,m, be framed in at least 60\% of the views and be visible by ray casting in at least half.
We drop an offset frame in which fewer than 200 pixels change by more than 8/255 inside the projected boxes of the object, which removes 21 of 5,123 frames.
We review the offset state of every arrangement visually and turn the defects into rejection rules: floating objects, contact with plants, hidden or unreadable objects, scenes that are too dark, holes under the support, and view-dependent render artifacts.
A rejected arrangement outside a too-dark scene is placed again where another pose passes, and 372 arrangements keep an offset state.

\noindent\textbf{Rendering and annotations.}
We render with Blender 5.1 EEVEE at 32 samples, or 16 in 72 scenes, depending on the rendering run.
Depth of field and motion blur are off, no lights are added, and the view transform is AgX.
Volumes can wash a frame out or be all that fills unlit space.
In 54 of the 66 scenes with volumes, we render a probe frame with and without them and keep the option with the higher score $\sigma-2v-1.5\max(0,\,\mu-0.62)$, where $\mu$ and $\sigma$ are the mean and standard deviation of its brightness and $v$ is its fraction of near-black pixels.
In total, 29 scenes lose their volumes, 10 of them without a probe, and 15 lose their compositor.
Layouts use a white background.
Ground-truth instance masks come from the same scene state with the Workbench engine, without anti-aliasing: a visible label map, and one amodal mask and one box mask per object.

\noindent\textbf{Boxes and annotations.}
The 3D box of an object is the bounding box of its mesh in the coordinate system of the asset, so it keeps its axes and face order in every state and view; objects stay upright, and an edit changes only the position and yaw of a box.
A layout image draws the boxes with translucent faces and solid edges, in one color per object that is fixed within a scene.
Each render comes with its layout image, 3D boxes, camera intrinsics and pose, and masks, and each asset with a category and captions.
The base, moved and added states carry an intersection flag, so that flagged arrangements can be filtered.
The release contains renders and annotations but no scene or asset files.

\noindent\textbf{Revisits and visibility.}
All states of an arrangement share their views, so an edit followed by its inverse returns to a rendered state and view, which gives an exact revisit; a path through the moved and offset states, or a return to a neighboring view, gives a revisit away from the earlier pose or view.
A revisit needs an earlier observation only when an edit changes what is visible.
In 306 of the 474 arrangements with valid visible masks, at least one view has such a change (\cref{tab:dataset}): the added object hides at least 10\% of another object, a movement changes the visible area of another object by at least 10\%, or the visible fraction of the moved object crosses 0.5.

\noindent\textbf{Limitations.}
Renders use the rasterizing EEVEE engine with the lights of the scene, so secondary effects depend on the scene: in 125 arrangements, moving the object changes no pixel farther than 4 pixels from it, so no shadow or reflection changes away from the object.
The visible label maps are empty in three scenes and for some objects in four more, and in six scenes with water, objects stand on the water surface.

\end{document}